\documentclass[letterpaper]{article} 
\usepackage{aaai24}  
\usepackage{times}  
\usepackage{helvet}  
\usepackage{courier}  
\usepackage[hyphens]{url}  
\usepackage{graphicx} 
\usepackage{natbib}  
\usepackage{caption} 
\usepackage{algorithm}
\usepackage{algorithmic}
\usepackage{newfloat}
\usepackage{amsmath}
\usepackage{subfigure}
\usepackage{multirow}
\usepackage{adjustbox}
\usepackage{amssymb}
\usepackage{bbding}
\usepackage{xcolor}

\usepackage{newfloat}
\usepackage{listings}
\DeclareCaptionStyle{ruled}{labelfont=normalfont,labelsep=colon,strut=off} 
\floatstyle{ruled}
\newfloat{listing}{tb}{lst}{}
\floatname{listing}{Listing}
\title{NuScenes-QA: A Multi-Modal Visual Question Answering Benchmark for Autonomous Driving Scenario}
\author{
    Tianwen Qian\textsuperscript{\rm 1},
    Jingjing Chen\textsuperscript{\rm 2}\equalcontrib,
    Linhai Zhuo\textsuperscript{\rm 2},
    Yang Jiao\textsuperscript{\rm 2},
    Yu-Gang Jiang\textsuperscript{\rm 2}
}
\affiliations{
    \textsuperscript{\rm 1}Academy for Engineering and Technology, Fudan University\\
    \textsuperscript{\rm 2}Shanghai Key Lab of Intelligent Information Processing, School of Computer Science, Fudan University\\
    \{twqian19, chenjingjing, lhzhuo19, ygj\}@fudan.edu.cn, yjiao23@m.fudan.edu.cn\\
}

\usepackage{bibentry}

\begin{document}

\maketitle

\begin{abstract}
We introduce a novel visual question answering (VQA) task in the context of autonomous driving, aiming to answer natural language questions based on street-view clues. Compared to traditional VQA tasks, VQA in autonomous driving scenario presents more challenges. Firstly, the raw visual data are multi-modal, including images and point clouds captured by camera and LiDAR, respectively. Secondly, the data are multi-frame due to the continuous, real-time acquisition. Thirdly, the outdoor scenes exhibit both moving foreground and static background. Existing VQA benchmarks fail to adequately address these complexities. To bridge this gap, we propose NuScenes-QA, the first benchmark for VQA in the autonomous driving scenario, encompassing 34K visual scenes and 460K question-answer pairs. Specifically, we leverage existing 3D detection annotations to generate scene graphs and design question templates manually. Subsequently, the question-answer pairs are generated programmatically based on these templates. Comprehensive statistics prove that our NuScenes-QA is a balanced large-scale benchmark with diverse question formats. Built upon it, we develop a series of baselines that employ advanced 3D detection and VQA techniques. Our extensive experiments highlight the challenges posed by this new task. Codes and dataset are available at \url{https://github.com/qiantianwen/NuScenes-QA}.
\end{abstract}

\section{Introduction}

\begin{figure}
    \centering
    \includegraphics[width=0.45\textwidth]{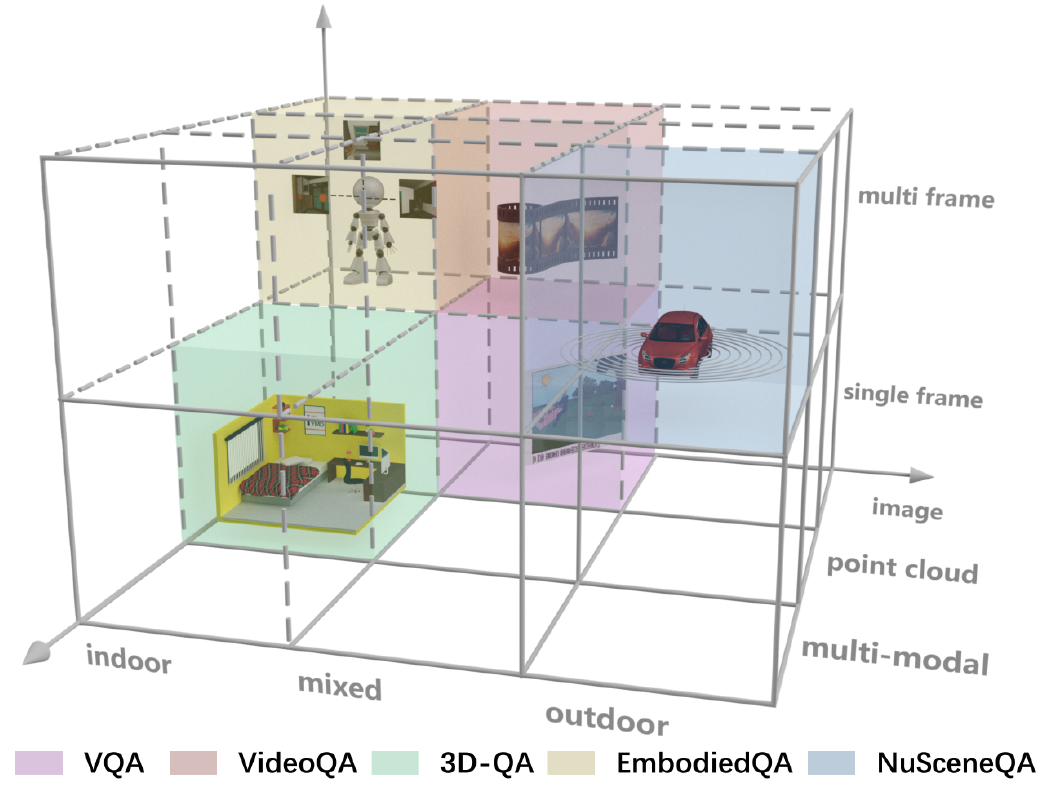}
    \caption{NuScenes-QA is a multi-modal, multi-frame, outdoor dataset that differs significantly from other VQA benchmarks in terms of visual data.}
    \label{fig:intro}
\end{figure}

Autonomous driving is a rapidly developing field with immense potential to improve transportation safety and efficiency with advancements in sensor technologies and computer vision. As the increasing maturity of traditional perceptual techniques such as 3D object detection \cite{liu2022bevfusion, jiao2023instance} and tracking \cite{chen2023voxenext}, autonomous driving systems are progressing towards enhanced interpretability and flexible human-car interactivity. In this context, visual question answering (VQA) \cite{antol2015vqa} can play a critical role. On one hand, VQA possesses interactive and entertainment, enabling passengers to perceive their surroundings through language and enhancing the user experience of intelligent driving systems. On the other hand, users can verify the correctness of perception system through question answering, fortifying their trust in its capabilities.

 Despite the notable progress made by the VQA community, models trained on existing VQA datasets \cite{balanced_vqa_v2, Hudson_2019_CVPR} have limitations in addressing the complexities of autonomous driving scenario. This limitation is primarily caused by the difference in visual data between self-driving scenario and existing VQA benchmarks.
 For instance, to answer question like ``\textit{Are there any moving pedestrians in front of the stopped bus?}'', it is necessary to locate and identify the bus, pedestrians, and their status accurately. This requires the model to effectively leverage the complementary information from images and point clouds to understand complex scenes and capture object dynamics from multiple frames of data streams. Therefore, it is essential to explore VQA in the context of multi-modal, multi-frame and outdoor scenario. However, existing VQA benchmarks cannot satisfy all these conditions simultaneously, as illustrated in Fig. \ref{fig:intro}. For instance, although 3D-QA \cite{azuma2022scanqa} and the self-driving scenario both focus on understanding the structure and spatial relationships of objects, 3D-QA is limited to single-modal (\emph{i.e.}, point cloud), single-frame, and static indoor scenes. The same goes for other benchmarks. To bridge this gap, we construct the first VQA benchmark specifically designed for autonomous driving scenario, named NuScenes-QA. NuScenes-QA is different from all other existing VQA benchmarks in terms of visual data characteristics, presenting new challenges for both VQA and autonomous driving community.

 The proposed NuScenes-QA is built upon nuScenes \cite{caesar2020nuscenes}, which is a popular 3D perception dataset for autonomous driving.
 We automatically annotate the question-answer pairs using the CLEVR benchmark \cite{johnson2017clevr} as inspiration. To be specific, we consider each keyframe annotated in nuScenes as a ``scene'' and construct a related scene graph. The objects and their attributes are regarded as the nodes in the graph, while the relative spatial relationships between objects are regarded as the edges, which are calculated based on the 3D bounding boxes annotated in nuScenes. Additionally, we design different types of question templates manually, including counting, comparison, and existence, etc.
 Based on these constructed templates and scene graphs, we sample different parameters to instantiate the templates, and use the scene graph to infer the correct answers, thus automatically generating question-answer pairs.
 Eventually, we obtained a total of 460K question-answer pairs on 34K scenes from the annotated nuScenes training and validation split, with 377K pairs for training and 83K for testing.

 In addition to the dataset, we also develop baseline models using the existing 3D perception \cite{huang2021bevdet, yin2021center, msmdfusion} and visual question answering \cite{anderson2018bottom, yu2019deep} techniques. These models fall into three categories: image-based, point cloud-based, and multi-modal fusion-based. The 3D detection models are used to extract visual features and provide object proposals, which are then combined with question features and fed into the question answering model for answer decoding. While our experiments show that these models outperform the question-only blind model, their performance still significantly lags behind models that use ground truth object labels as inputs. This indicates that combining existing technologies is not sufficient for intricate street views understanding.
 Thus, NuScenes-QA poses a new challenge, urging further research in this realm.

 Overall, our contributions can be summarized as follows:
 \begin{itemize}
 \item We introduce a novel visual question answering task in autonomous driving scenario, which evaluates current deep learning based models' ability to understand and reason with complex visual data in multi-modal, multi-frame, and outdoor scenes. To facilitate this task, we contribute a large-scale dataset, NuScenes-QA, consisting of 34K complex autonomous driving scenes and 460K question-answer pairs. 
 \item We establish several baseline models and extensively evaluate the performance of existing techniques for this task. Additionally, we conduct ablation experiments to analyze specific techniques that are relevant to this task, which provide a foundation for future research. 
 \end{itemize}

 \begin{figure*}[ht]
    \centering
    \includegraphics[width=\textwidth]{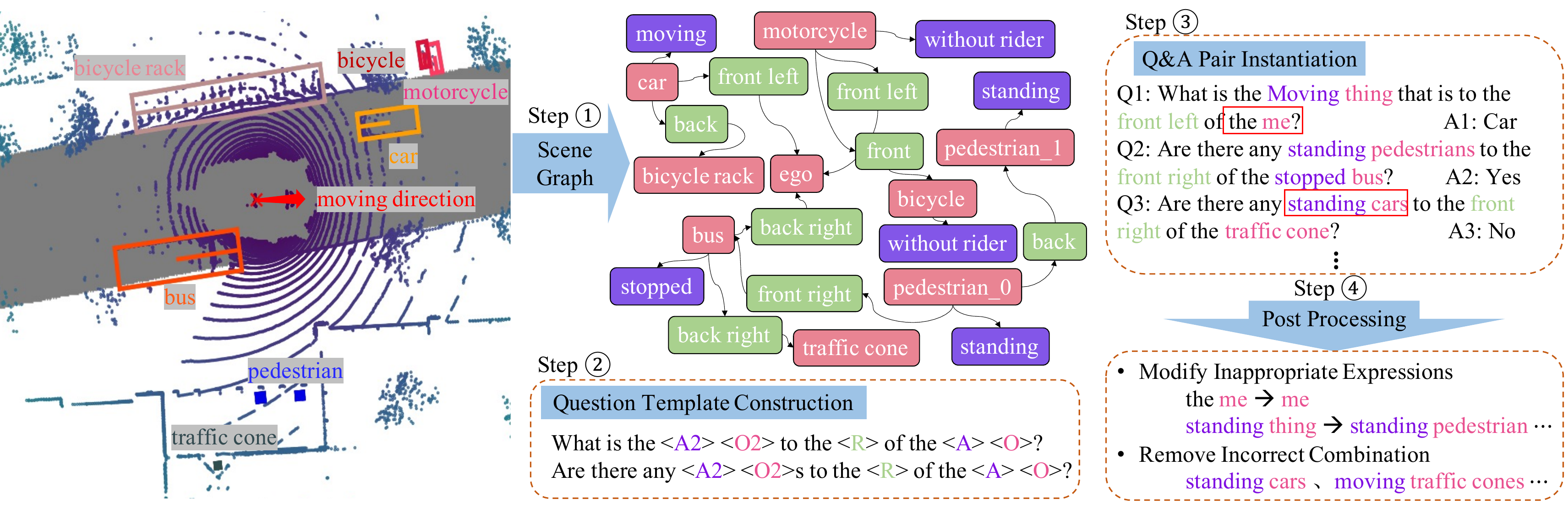}
    \caption{Data construction flow of NuScenes-QA. First, the scene graphs are generated using the annotated object labels and 3D bounding boxes. Then, we design question templates manually, and instantiate the question-answer pairs with them. Finally, the generated data are filtered based on certain rules.}
    \label{fig:data}
\end{figure*}

\section{Related Works}
\subsection{Visual Question Answering}
There are various datasets available for VQA, including image-based datasets such as VQA2.0 \cite{balanced_vqa_v2}, CLEVR \cite{johnson2017clevr}, and GQA \cite{Hudson_2019_CVPR}, as well as video-based datasets such as TGIF-QA \cite{jang2017tgif} and TVQA \cite{lei2018tvqa}. For the image-based VQA, earlier works \cite{lu2016hierarchical, anderson2018bottom, qian2022scene} typically use CNNs to extract image features, and RNNs to process the question. Then, joint embddings of vision and language obtained through concatenation or other operations \cite{kim2018bilinear} are input to the decoder for answer prediction. Recently, many Transformer-based models \cite{tan2019lxmert, zhang2021vinvl} have achieved state-of-the-art performance through large-scale vision-language pre-training. Differing to image-based VQA, VideoQA \cite{jang2017tgif, zhu2017uncovering} places greater emphasis on mining the temporal contextual from videos. For example, Jiang \textit{et al.} \cite{jiang2020divide} proposed a question-guided spatial-temporal contextual attention network, and Qian \textit{et al.} \cite{qian2022locate} suggested first localizing relevant segments in a long-term video before answering.

\subsection{3D Visual Question Answering}
3D Visual Question Answering (3D-QA) is a novel task in the VQA field that focuses on answering questions about 3D scenes represented by point cloud.
Unlike traditional VQA tasks, 3D-QA requires models to understand the geometric structure and the spatial relations of objects in a indoor scene. Recently, many 3D-QA datasets have been constructed. For example, the 3DQA dataset \cite{ye20223d}, which is based on ScanNet \cite{dai2017scannet}, has manually annotated 6K question-answer pairs. Similarly, ScanQA \cite{azuma2022scanqa} has utilized a question generation model along with manual editing to annotate 41K pairs on the same visual data.
Despite these advancements, current 3D-QA models face limitations in solving more complex autonomous driving scenario, which involve multi-modalities, multi-frames, and outdoor scenes.

\subsection{Vision-Language Tasks in Autonomous Driving}
Language systems are pivotal for communication between the passengers and vehicles. Pioneering efforts have explored language-guided visual understanding in this context. For instance, Deruyttere \textit{et al.} proposed the Talk2Car \cite{deruyttere2019talk2car}, which is the first object referral dataset with natural language commands for self-driving cars.
Wu \textit{et al.} developed a benchmark with scalable expressions named Refer-KITTI \cite{wu2023referring} based on the self-driving dataset KITTI \cite{geiger2012we}. It aims to track multiple targets based on language descriptions. In contrast, our proposed NuScene-QA stands out in two ways. Firstly, it tackles high-level question answering, demanding both understanding and reasoning. Secondly, NuScenes-QA provides richer visual information, including images and point clouds.

\section{NuScenes-QA Dataset}
Our primary contribution is the NuScenes-QA dataset, which we will introduce in detail in this section. We provide a comprehensive overview of the dataset construction, including scene graph development, question template design, question-answer pair generation, and post-processing. In addition, we analyze the statistical characteristics of the NuScenes-QA dataset, such as the distribution of question types, lengths, and answers.

\subsection{Data Construction}
\label{data_cons}

For question-answer pairs generation, we adapted an automated method inspired by CLEVR \cite{johnson2017clevr}. This method requires two types of structured data: scene graphs generated from 3D annotations, containing object category, position, and relationships; alongside manually crafted question templates that specify the question type, expected answer type, and reasoning required to answer it. By combining these structured data, we automatically generate question-answer pairs. These pairs are then filtered and validated through post-processing programs to construct the complete dataset. Fig. \ref{fig:data} illustrates the overall data construction pipeline.

\begin{figure*}[ht]
    \centering
    \subfigure[Question Length Distribution]{\includegraphics[width=0.3\textwidth]{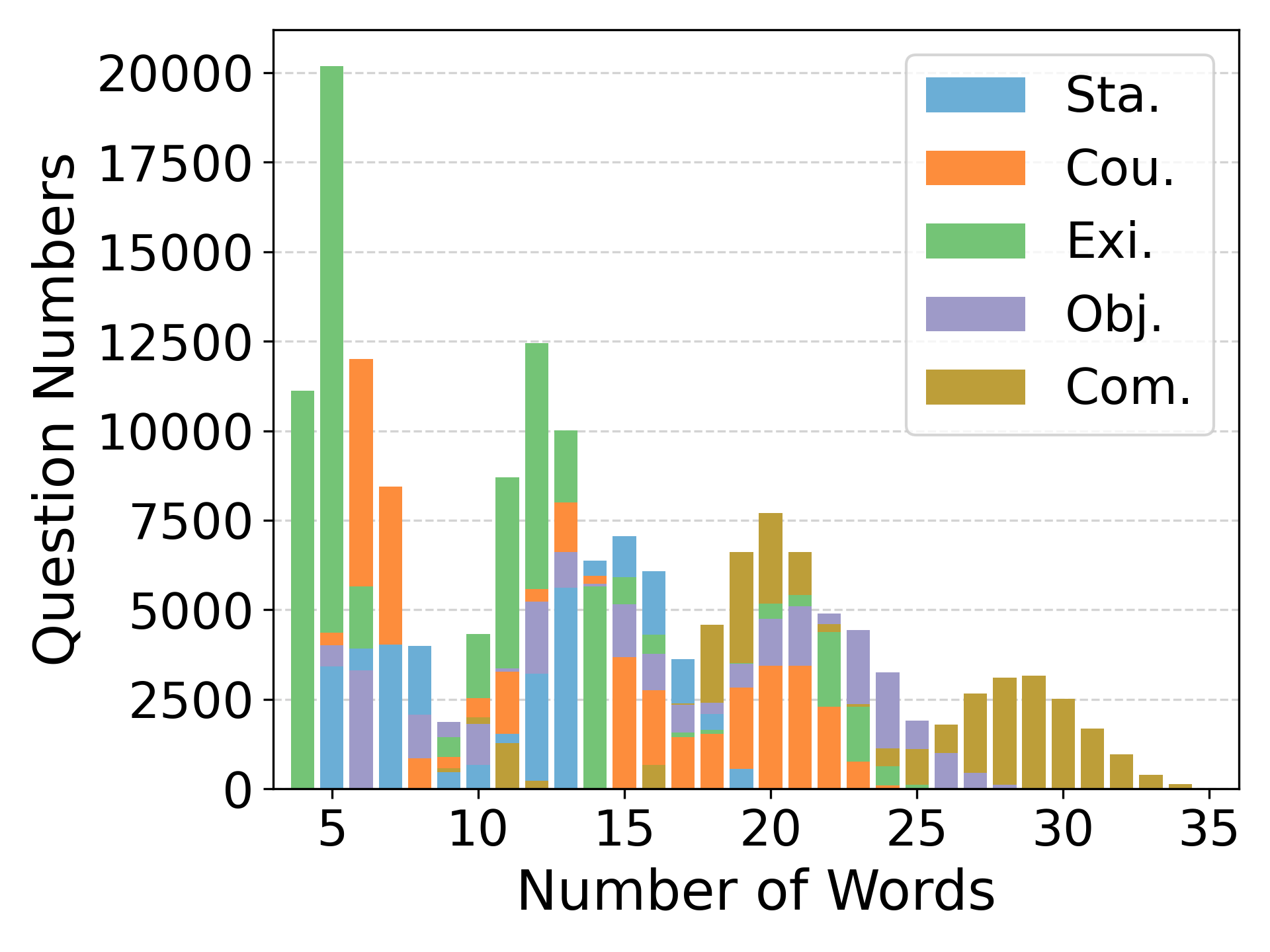}}
    \hspace{3mm}
    \subfigure[Answer Distribution]{\includegraphics[width=0.3\textwidth, height=0.22\textwidth]{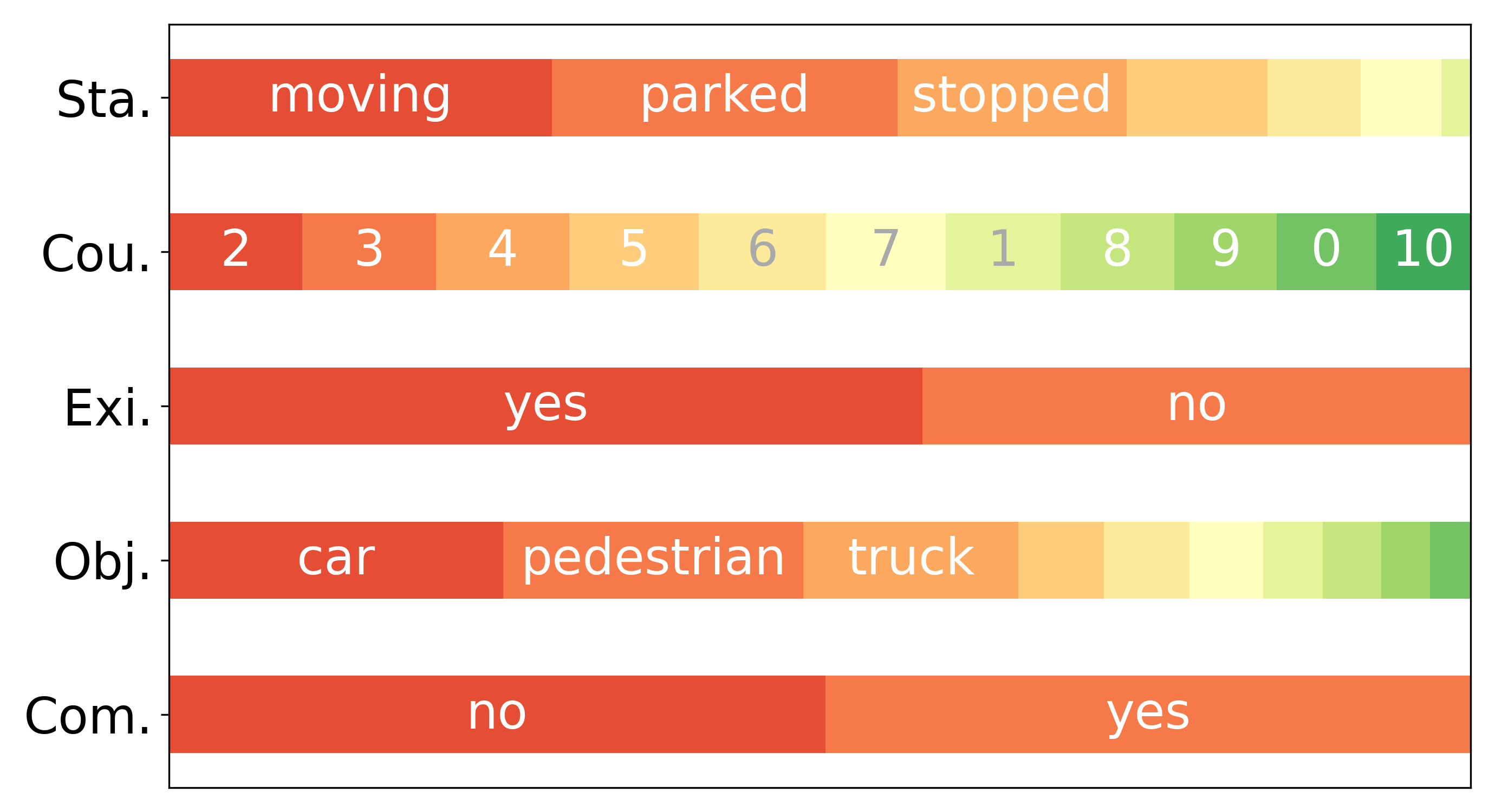}}
    \hspace{1mm}
    \subfigure[Category Distribution]{\includegraphics[width=0.3\textwidth]{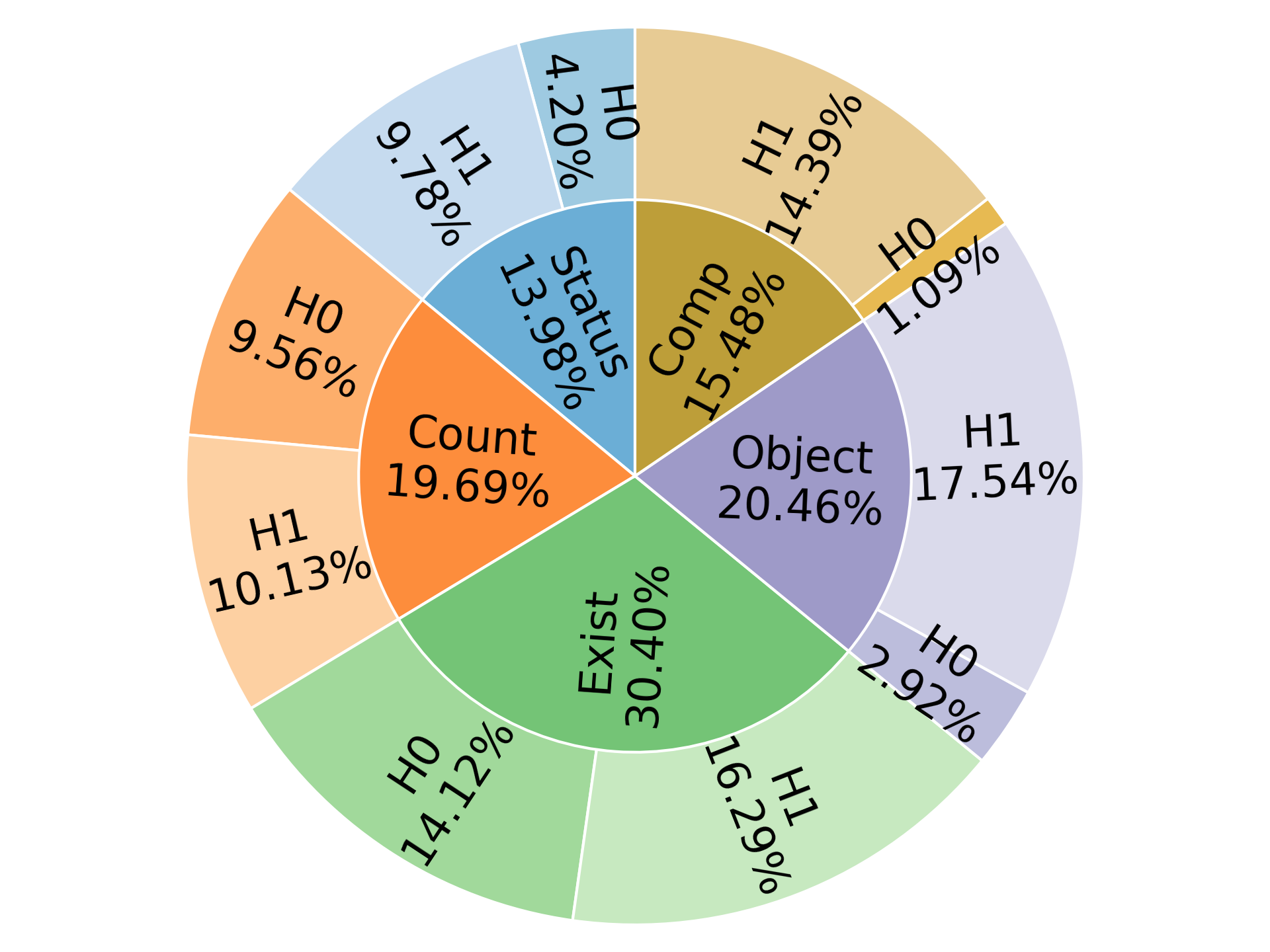}}
    \caption{Statistical distributions of questions and answers in the NuScenes-QA training split.}
    \label{fig:status}
\end{figure*}

\paragraph{\textbf{Scene Graph Construction}}
A scene graph \cite{johnson2015image} is defined as an abstract representation of a visual scene, where nodes in the graph represent objects in the scene, and edges represent relationships between objects. 

In nuScenes, the collected data is annotated with a frequency of 2Hz, and each annotated frame is referred as a ``keyframe''. We consider each keyframe as a ``scene'' in NuScenes-QA. The existing annotations include object categories and their attributes in the scene, as well as the 3D bounding boxes of the objects. These annotated objects and their attributes are directly used as nodes in the graph.
However, relationships between objects are not provided in the original annotations, so we developed a rule for calculating object relationships. Given that spatial position relationships are crucial in autonomous driving scenario, we define six relationships between objects, namely \textit{front}, \textit{back}, \textit{front left}, \textit{front right}, \textit{back left}, and \textit{back right}. To determine object relationships, we first project 3D bounding boxes onto the Bird's-Eye-View (BEV). Subsequently, we calculate the angle between the vector connecting the centers of two bounding boxes and the forward direction of the ego-car. The formula is given by
\begin{equation}
    \theta=\cos^{-1}\frac{(B_1[:2]-B_2[:2])\cdot V_{ego}[:2]}{\Vert B_1[:2]-B_2[:2] \Vert  \Vert  V_{ego}[:2] \Vert}, 
\end{equation}
where $B_i=[x, y, z, x_{size}, y_{size}, z_{size}, \varphi]$ is the 3D bounding box of object $i$, and $V_{ego}=[v_x, v_y, v_z]$ represents the speed of the ego car. Based on the angle range, the relationship between two objects is defined as
\begin{equation}
 relation=\begin{cases}
  front& \text{ if } -30^{\circ} < \theta <= 30^{\circ} \\
  front \ left& \text{ if } 30^{\circ} < \theta <= 90^{\circ} \\
  front \ right& \text{ if } -90^{\circ}<\theta<=-30^{\circ} \\
  back \ left& \text{ if } 90^{\circ}<\theta<=150^{\circ} \\
  back \ rigth& \text{ if } -150^{\circ}<\theta<=-90^{\circ} \\
  back& \text{ else. } 
\end{cases}
\end{equation}
We define the forward direction of the car as $0^{\circ}$ and counterclockwise angle as positive. At this point, we can convert the annotations of nuScenes into the scene graphs we need, as illustrated in step one of Fig. \ref{fig:data}.

\paragraph{\textbf{Question Template Design}}

We devised templates manually for question generation. For instance, the question ``\textit{What is the moving thing to the front left of the stopped bus?}'' can be abstracted as the template ``\textit{What is the \textless A2\textgreater \textless O2\textgreater to the \textless R\textgreater of the \textless A1\textgreater \textless O1\textgreater?}'', with \textit{\textless A\textgreater}, \textit{\textless O\textgreater}, and \textit{\textless R\textgreater} as parameters for instantiation, representing attribute, object, and relationship, respectively. Additionally, we can express the same semantic with another form like ``\textit{There is a \textless A2\textgreater \textless O2\textgreater to the \textless R\textgreater of the \textless A1\textgreater \textless O1\textgreater, what is it?}''.

Ultimately, NuScenes-QA holds 66 diverse question templates, divided into 5 question types: \textit{existence}, \textit{counting}, \textit{query-object}, \textit{query-status}, and \textit{comparison}. In addition, to better evaluate the models reasoning performance, we also divide the questions into \textit{zero-hop} and \textit{one-hop}. Specifically, zero-hop questions require no reasoning between objects, \emph{e.g.}, ``\textit{What is the status of the \textless A\textgreater \textless O\textgreater?}''. One-hop questions involve one step spatial reasoning, \emph{e.g.}, ``\textit{What is the status of the \textless A2\textgreater \textless O2\textgreater to the \textless R\textgreater of the \textless A1\textgreater \textless O1\textgreater?}''. Comprehensive template details are available in the supplementary material.

\paragraph{\textbf{Q\&A Pair Generation and Filtering}}

Given the scene graphs and question templates, instantiating a question-answer pair is straightforward: we select a template, assign parameter values through depth-first search, and deduce the ground truth answer on the scene graph. Moreover, we dismiss ill-posed or degenerate questions. For instance, the question is ill-posed if the scene do not contain any cars or pedestrians when \textit{\textless O1\textgreater==pedestrian} and \textit{\textless O2\textgreater==car} is assigned for the template ``\textit{What is the status of the \textless O2\textgreater to the \textless R\textgreater of the \textless A1\textgreater \textless O1\textgreater?}''.

It is important to note that post-processing, as depicted in step 4 of Fig. \ref{fig:data}, addresses numerous unsuitable expressions. For example, we added the ego-car as an object in the scene, it is referred to as ``\textit{me}'' in questions. This led to some inappropriate instances like ``\textit{the me}'' or ``\textit{there is a me}'' when \textit{\textless O\textgreater} is assigned ``\textit{me}''. We revised such expressions. In addition, during the instantiation, inappropriate \textit{\textless A\textgreater+\textless O\textgreater} combinations like ``\textit{standing cars}'' and ``\textit{parked pedestrians}'' were eliminated through rules. Also, we removed questions with counting answers greater than 10 to balance the answer distribution.


\begin{figure*}[t]
    \centering
    \includegraphics[width=\textwidth]{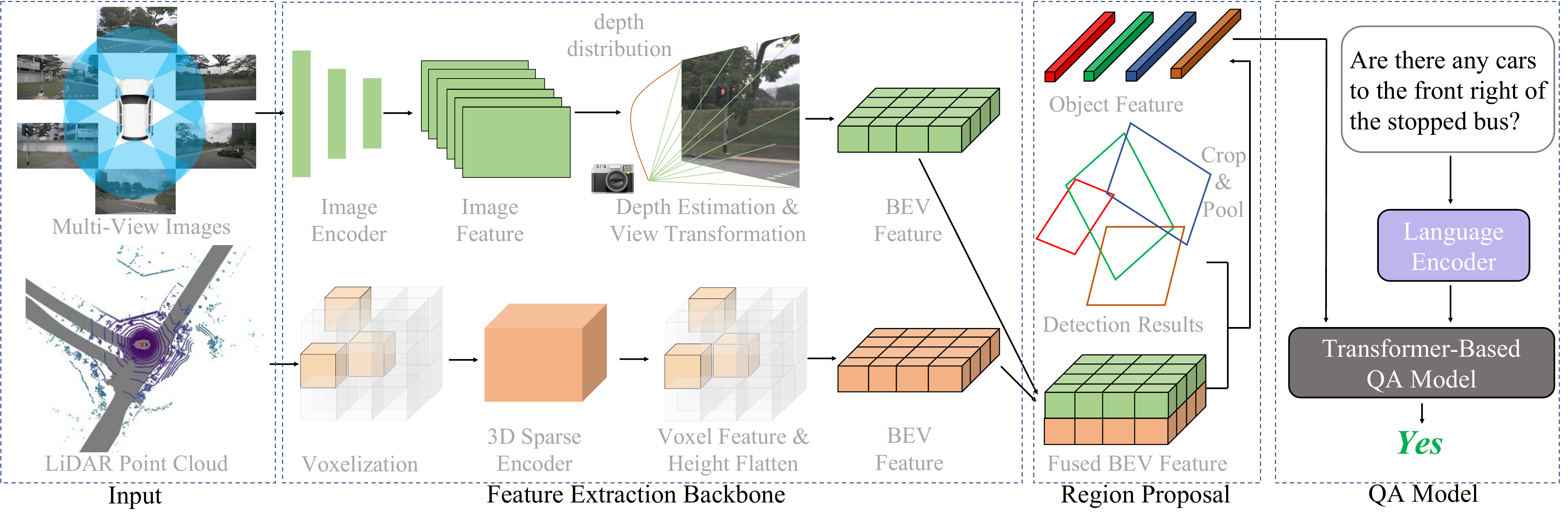}
    \caption{Framework of baseline. The multi-view images and point clouds are first processed by the feature extraction backbone to obtain BEV features. Then, the objects embeddings are cropped based on the detected 3D bounding boxes. Finally, these objects features are fed into the question-answering head along with the given question for answer decoding.}
    \label{fig:framework}
\end{figure*}

\subsection{Data Statistics}
\label{data_sta}

In total, NuScenes-QA provides 459,941 question-answer pairs across 34,149 visual scenes, with 376,604 questions from 28,130 scenes for training, and 83,337 questions from 6,019 scenes for testing. To the best of our knowledge, NuScenes-QA is currently the largest 3D related question answering dataset. Detailed comparison of 3D-QA datasets can be found in supplementary materials.


Fig. \ref{fig:status} depicts various statistical distributions of NuScenes-QA.
Fig. \ref{fig:status}(a) showcases a broad spectrum of question lengths (5 to 35 words). Fig. \ref{fig:status}(b) and \ref{fig:status}(c) present answer and question category distributions, revealing the balance of NuScenes-QA. A balanced dataset can prevent models from learning answer biases or language shortcuts, which are common in many other VQA benchmarks \cite{antol2015vqa, azuma2022scanqa}.



\section{Method}

Along with the proposed dataset, we provide several baselines based on existing 3D detection and VQA techniques.

\subsection{Task Definition}
\label{sec:task}

Given a visual scene $S$, and a question $Q$, the task of visual question answering aims to select an answer $\hat{a}$ from the answer space $\mathcal{A}=\{a_i\}_{i=1}^{N}$ that best answers the question. Therefore, the task can be formulated as:
\begin{equation}
    \hat{a}=\mathop{\arg\max}\limits_{a\in \mathcal{A}}P(a\mid S,Q).
    \label{eq:vqa}
\end{equation}
For NuScenes-QA, visual scene data encompass multi-view images $I$, point clouds $P$, and any frames $I_i$ and $P_i$ before the current frame in the data sequences. We can further decompose the Eq. \ref{eq:vqa} into:
\begin{align}
    &P(a\mid S,Q)=P(a\mid I,P,Q)
     \notag
     \\
    &I=\{I_i,\ T-t<i\le T\}
    \\
    &P=\{ P_i,\ T-t<i\le T\},
    \notag
\end{align}
where $T$ is the index of current frame and $t$ is the number of previous frame used in the model. It is also possible to use only single modality or single frame data for prediction.

\subsection{Framework Overview}
\label{sec:framework}

The overall framework of our proposed baseline is illustrated in Fig. \ref{fig:framework} and mainly consists of three components. The first is the feature extraction backbone, which includes both image and point cloud feature extractor. The second part is the region proposal module for object embedding, and the last component is the QA-head for answer prediction.

Initially, the surrounded-view images and point clouds are fed into the feature extraction backbone, with features projected onto the Bird's-Eye-View (BEV). Subsequently, 3D bounding boxes inferred by a pre-trained detection model are used to crop and pool object features. Finally, the QA-model takes the question features and the object features as input for cross-modal interaction to predict the answer.

\subsection{Input Embedding}
\label{sec:embed}

\paragraph{\textbf{Question Embedding}}
For a question $Q=\{w_i\}_{i=1}^{n_q}$ that contains $n_q$ words, we first tokenize it and initialize the tokens with pre-trained GloVe \cite{pennington2014glove} embeddings. The sequence is then fed into a single-layer biLSTM \cite{hochreiter1997long} for word-level context encoding. Each word feature $\mathbf{w}_i$ is represented by the concatenation of the forward and backward hidden states of the biLSTM, denoted as:
\begin{equation}
    \mathbf{w}_i = [\overset{\rightarrow}{\mathbf{h}_i};\ \overset{\leftarrow}{\mathbf{h}_i}]\in \mathbb{R}^{d},
\end{equation}
and the question embedding is represented as $\mathbf{Q}\in \mathbb{R}^{n_q\times d}$.

\paragraph{\textbf{Visual Feature Extraction}}

We adopt leading-edge 3D detection techniques for visual feature extraction. As shown in Fig. \ref{fig:framework}, it entails two branches: image stream and point cloud stream. For multi-view images, ResNet \cite{he2016deep} with FPN \cite{lin2017feature} is used as the backbone for multi-scale feature extraction. Then, in order to make the feature spatial-aware, we estimate the depth of the pixels in the images and lift them to 3D virtual points with a view transformer inspired by LSS \cite{philion2020lift}. Finally, pooling along the Z-axis compresses the feature in voxel space, producing the BEV featmap $\mathbf{M}_I\in  \mathbb{R}^{H\times W\times d_m} $. 

For point clouds, we first partition 3D space into voxels, transforming raw point clouds into binary voxel grids \cite{zhou2018voxelnet}. Subsequently, 3D sparse convolutional neural network \cite{graham20183d} is applied to the voxel grid for feature representation. Similar to the image features mentioned earlier, Z-axis pooling yields the point cloud BEV featmap $\mathbf{M}_P\in  \mathbb{R}^{H\times W\times d_m}$. Combining $\mathbf{M}_I$ and $\mathbf{M}_P$, we can aggregate them to obtain multi-modal featmap $\mathbf{M}\in  \mathbb{R}^{H\times W\times d_m}$.

\paragraph{\textbf{Object Embedding}}

Following 2D detection works \cite{ren2015faster}, we crop and pool the features in bounding boxes as the object embedding. However, unlike standard 2D bounding boxes aligned with the coordinate axis in images, projecting 3D boxes to BEV yields rotated boxes unsuited for standard RoI Pooling. To this end, we make some modifications.
Firstly, we project the 3D box $B=[x, y, z, x_{size}, y_{size}, z_{size}, \varphi]$ into the BEV featmap:
\begin{equation}
    \label{eq:trans}
    x_m = \frac{x-R_{pc}}{F_v\times F_o},
\end{equation}
where,$F_v$, $F_o$ and $R_{pc}$ represent the voxel factor, out size factor of the backbone, and the point cloud range, respectively. All box parameters follow the Eq. \ref{eq:trans} to transform into BEV space except the heading angle $\varphi$. Then, based on the center and size of the box, we can easily calculate the four vertices $V=\{x_i, y_i\}_{i=0}^{3}$. Secondly, we calculate the rotated vertex $V^{'}$ via the heading angle $\varphi$:
\begin{equation}
    \begin{bmatrix}
 x_i^{'}\\
y_i^{'}
\end{bmatrix}=\begin{bmatrix}
 \cos\varphi & -\sin\varphi \\
 \sin\varphi & \cos\varphi
\end{bmatrix}\begin{bmatrix}
 x_i\\
y_i
\end{bmatrix}
\end{equation}
Finally, we use cross product algorithm to identify pixel membership in the rotated rectangle. Then, we perform mean pooling on the features of all the pixels inside the rectangle to obtain the object embedding $\mathbf{O}\in \mathbb{R}^{N\times d_m} $. Algorithm details can be found in supplementary materials.





\subsection{Answer Head and Training}
\label{sec:head}

We adopt the classical VQA model MCAN \cite{yu2019deep} as our answer head. It leverages stacked self-attention layers to model the language and visual context independently, along with stacked cross-attention layers for cross-modal feature interaction. The fused features are then projected to the answer space for prediction via basic MLP layers.

During the training phase, we extract the object embeddings using a pre-trained 3D detection model offline. And answer head is trained with the standard cross-entropy loss.

\begin{table*}[ht]
\centering
\begin{adjustbox}{width=\textwidth}
\begin{tabular}{c|ccc|ccc|ccc|ccc|ccc|c}
\hline
\multirow{2}{*}{Models} & \multicolumn{3}{c|}{Exist} & \multicolumn{3}{c|}{Count} & \multicolumn{3}{c|}{Object} & \multicolumn{3}{c|}{Status} & \multicolumn{3}{c|}{Comparison} & \multirow{2}{*}{Acc} \\ \cline{2-16}
& H0 & H1 & All & H0 & H1 & All & H0 & H1 & All & H0 & H1 & All & H0 & H1 & All & \\ \hline
Q-Only & 81.7 & 77.9 & 79.6 & 17.8 & 16.5 & 17.2 & 59.4 & 38.9 & 42.0 & 57.2 & 48.3 & 51.3 & 79.5 & 65.7 & 66.9 & 53.4 \\
\hline
\hline
BEVDet+BUTD & 87.2 & 80.6 & 83.7 & 21.7 & 20.0 & 20.9 & 69.4 & 45.2 & 48.8 & 55.0 & 50.5 & 52.0 & 76.1 & 66.8 & 67.7 & 57.0 \\
CenterPoint+BUTD & 87.7 & 81.1 & 84.1 & 21.9 & 20.7 & 21.3 & 70.2 & 45.6 & 49.2 & 62.8 & 52.4 & 55.9 & 81.6 & 68.0 & 69.2 & 58.1 \\
MSMDFusion+BUTD & 89.4 & 81.4 & 85.1 & 25.3 & 21.3 & 23.2 & 73.3 & 48.7 & 52.3 & 67.4 & 55.4 & 59.5 & 81.6 & 67.2 & 68.5 & 59.8 \\
GroundTruth+BUTD & 98.9 & 87.2 & 92.6 & 76.8 & 38.7 & 57.5 & 99.7 & 71.9 & 76.0 & 98.8 & 81.9 & 87.6 & 98.1 & 76.1 & 78.1 & 79.2 \\
\hline
\hline
BEVDet+MCAN & 87.2 & 81.7 & 84.2 & 21.8 & 19.2 & 20.4 & 73.0 & 47.4 & 51.2 & 64.1 & 49.9 & 54.7 & 75.1 & 66.7 & 67.4 & 57.9 \\
CenterPoint+MCAN & 87.7 & 82.3 & 84.8 & 22.5 & 19.1 & 20.8 & 71.3 & 49.0 & 52.3 & 66.6 & 56.3 & 59.8 & 82.4 & 68.8 & 70.0 & 59.5 \\
MSMDFusion+MCAN & 89.0 & 82.3 & 85.4 & 23.4 & 21.1 & 22.2 & 75.3 & 50.6 & 54.3 & 69.0 & 56.2 & 60.6 & 78.8 & 68.8 & 69.7 & 60.4 \\ 
GroundTruth+MCAN & 99.6 & 95.5 & 97.4 & 52.7 & 39.9 & 46.2 & 99.7 & 86.2 & 88.2 & 99.3 & 95.4 & 96.8 & 99.7 & 90.2 & 91.0 & 84.3 \\ 
\hline
\end{tabular}
\end{adjustbox}
\caption{Results of different models on the NuScenes-QA test set. We evaluate top-1 accuracy across the overall test split and different question types. H0 denotes zero-hop and H1 denotes one-hop. C denotes camera, L denotes LiDAR.}
\label{tab:sota}
\end{table*}

\section{Experiments}

To validate the challenge of NuScenes-QA, we assess baseline performance in various configurations: camera-only or LiDAR-only single-modality models, camera-lidar fusion models, and diverse answering heads. We conduct ablation studies on crucial steps of the baseline, including BEV feature cropping and pooling strategies, as well as the influence of detected 3D bounding boxes. Furthermore, visualization samples are showcased in supplementary material.

\subsection{Evaluation Metrics}
Questions in NuScenes-QA span 5 categories based on query format: 1) \textbf{Exist}, querying the existence of a object in the scene; 2) \textbf{Count}, object counting under specified conditions; 3) \textbf{Object}, object recognition based on language description; 4) \textbf{Status}, querying the status of a specified object; 5) \textbf{Comparison}, specified objects or their status comparison. Additionally, questions are also divided into two groups based their complexity of reasoning: zero-hop (denoted as \textbf{H0}) and one-hop (denoted as \textbf{H1}). We adopt the Top-1 accuracy as our evaluation metric, follow the practice of many other VQA works \cite{antol2015vqa, azuma2022scanqa}, and evaluate the performance of different question types separately.

\subsection{Implementation Details}

For the feature extraction backbone, we use the pre-trained detection model following the original settings \cite{huang2021bevdet, yin2021center, msmdfusion}.
The dimension of the QA model $d_m$ is set to 512, and MCAN adopts a 6-layer encoder-decoder version. As for training, we used the Adam optimizer with an initial learning rate of 1e-4 and half decaying every 2 epochs. All experiments are conducted with a batch size of 256 on 2 NVIDIA GeForce RTX 3090 GPUs. More details can be found in supplementary material.

\subsection{Quantitative Results}
\paragraph{\textbf{Compared Methods}}

As mentioned earlier, our task can be divided into three settings: camera-only, LiDAR-only, camera+LiDAR. To explore the impact of different modalities on the question-answering performance, we select representative backbone for each setting. We choose \textbf{BEVDet} \cite{huang2021bevdet} for camera-only setting, which proposed a novel paradigm of explicitly encoding the perspective-view features into the BEV space. \textbf{CenterPoint} \cite{yin2021center} is selected for LiDAR-only setting. It introduced a center-based object keypoint detector and has shown excellent performance in both detection accuracy and speed. For the multi-modal model, we opt for \textbf{MSMDFusion} \cite{msmdfusion}, which leverages depth and fine-grained LiDAR-camera interaction, achieving state-of-the-art results on the nuScenes detection benchmark for single model.

Regarding the QA-head, we select two classic models, \textbf{BUTD} \cite{anderson2018bottom} and \textbf{MCAN} \cite{yu2019deep}. BUTD advocates for computing bottom-up and top-down attention on salient regions of the image. MCAN stacks self-attention and cross-attention modules for vision-language feature interaction. To establish the upper bound of the QA models, we employ perfect perceptual results, \emph{i.e.}, ground-truth object labels. Specifically, we use GloVe for objects and their status embedding, noted as \textbf{GroundTruth} in Table \ref{tab:sota}. Additionally, we design a \textbf{Q-Only} baseline to investigate the impact of language bias. Q-Only can be considered as a blind model that ignores the visual input.

\paragraph{\textbf{Results and Discussions}}
According to the results shown in Table \ref{tab:sota}, we have the following observations that are worth discussing.

1. It is evident that visual data play a critical role in the performance of our task. When comparing the Q-Only baseline to others, we find that it only achieves an accuracy of $53.4\%$, which is significantly lower than that of other models. For instance, MSMDFusion+MCAN performs $7\%$ better. This indicates that model cannot achieve good performance solely rely on language shortcuts, but needs to leverage rich visual information.

2. Referring to the bottom part of Table \ref{tab:sota}, we can see that the LiDAR-based CenterPoint outperforms the camera-based BEVDet, achieving accuracy of $57.9\%$ and $59.5\%$, respectively. We attribute this performance gap to the task characteristics. Images possess detailed texture information, point clouds excel in structure and spatial representation. Our proposed NuScenes-QA emphasizes more on the understanding of structure and spatial relationships of objects. On the other hand, the fusion-based model MSMDFusion attains the best performance with an accuracy of $60.4\%$, demonstrating the camera and LiDAR data are complementary. Further work can explore how to better exploit the complementary information of multi-modal data. Of course, our baselines still have a long way to go compared to the GroundTruth (achieving an accuracy of $84.3\%$).

3. According to Table \ref{tab:sota}, QA-head has a significant impact on the performance. With the same detection backbone, we observed that the QA-head based on MCAN outperforms BUTD by a large margin. For example, the overall accuracy of CenterPoint+MCAN is $59.5\%$, $1.4\%$ higher than CenterPoint+BUTD. A dedicated QA-head designed for NuScenes-QA may lead to a greater improvement. We leave this as future work.

4. In a horizontal comparison of Table \ref{tab:sota}, it is not difficult to find that counting is the most difficult among all question types. Our best baseline model achieved just $23.2\%$ accuracy, much lower than other question types. Counting is historically tough in visual question answering, and some explorations \cite{zhang2018learning} have been made in traditional 2D-QA. Future efforts could involve counting modules to enhance its performance.

\subsection{Ablation Studies}

\begin{table}[]
\begin{adjustbox}{width=0.45\textwidth}
\begin{tabular}{c|ccccc|c}
\hline
\multirow{2}{*}{Variants} & \multicolumn{5}{c|}{Question Types} & \multirow{2}{*}{All} \\
& Exi. & Cou. & Obj. & Sta. & Com. & \\ \hline
Det w/o boxes & 84.8 & 20.8 & 52.3 & 59.8 & 70.0 & 59.5 \\
Det w/ boxes & 84.3 & 21.7 & 53.0 & 57.7 & 67.2 & 58.9 \\ 
\hline
\hline
GT w/o boxes & 91.2 & 38.8 & 61.1 & 80.3 & 76.8 & 70.8 \\
GT w/ boxes & 97.4 & 46.2 & 88.2 & 96.8 & 91.0 & 84.3 \\ 
\hline
\end{tabular}
\end{adjustbox}
\caption{Ablation comparison between model trained with and without detection boxes feature.}
\label{tab:box}
\end{table}

To validate effectiveness of different operations in our baselines, we conduct extensive ablation experiments on the NuScenes-QA test split using the CenterPoint+MCAN baseline combination. 

\paragraph{\textbf{Effects of Bounding Boxes}}
Most 2D and 3D VQA models fuse the visual feature with object bounding box in the object embedding stage, making it position-aware. We follow this paradigm and evaluate the impact of 3D bounding boxes in our NuScenes-QA. Specifically, we project the 7-dimensional box $B=[x, y, z, x_{size}, y_{size}, z_{size}, \varphi]$ obtained from the detection model onto the same dimension as the object embeddings using MLP, and concatenate the two features as the final input for the QA head. As shown in Table \ref{tab:box}, we are surprised to find that the performance varies significantly on different data. Adding box features for ground truth can increase the model's accuracy from $70.8\%$ to $84.3\%$, a significant improvement of $13.5\%$. However, adding the detected boxes slightly decreased performance by $0.6\%$, which is counterproductive. We speculate that this phenomenon may be caused by two reasons. On one hand, the current 3D detection models are still immature, and the noise in the detected boxes hurts the QA model. On the other hand, the point cloud represented by XYZ itself has great position expression ability, and the gain from adding box features is not significant.

\paragraph{\textbf{BEV Feature Crop Strategy}}
As mentioned earlier, due to the non-parallelism between the 3D boxes and the BEV coordinate axes, we cannot perform standard RoI pooling as in traditional 2D images. Therefore, we use cross product algorithm to determine pixels inside the rotated box for feature cropping. In addition to this method, we can also use a simpler approach, which directly uses the circumscribed rectangle of the rotated box parallel to the coordinate axes as the cropping region. Table \ref{tab:crop} shows the performance comparison of these two crop strategy, where the circumscribed box is slightly inferior to the rotated box. The reason for this is that NuScenes-QA contains many elongated objects, such as bus and truck. These objects occupy a small area in the BEV space, but their circumscribed rectangles have a large range, making the object features over smoothing.

\begin{table}[]
\centering
\begin{adjustbox}{width=0.45\textwidth}
\begin{tabular}{c|ccccc|c}
\hline
\multirow{2}{*}{Crop} & \multicolumn{5}{c|}{Question Types} & \multirow{2}{*}{All} \\
& Exi. & Cou. & Obj. & Sta. & Com. & \\ \hline
Cir. Box & 84.0 & 21.8 & 52.3 & 60.2 & 65.2 & 58.8 \\
Rot. Box & 84.8 & 20.8 & 52.3 & 59.8 & 70.0 & 59.5 \\ \hline
\end{tabular}
\end{adjustbox}
\caption{Ablation comparison of BEV feature crop strategies.}
\label{tab:crop}
\end{table}

\begin{table}[]
\centering
\begin{adjustbox}{width=0.45\textwidth}
\begin{tabular}{c|ccccc|c}
\hline
\multirow{2}{*}{Pooling} & \multicolumn{5}{c|}{Question Types} & \multirow{2}{*}{All} \\
& Exi. & Cou. & Obj. & Sta. & Com. & \\ \hline
Max & 84.2 & 20.7 & 51.6 & 58.0 & 69.7 & 58.9 \\
Mean & 84.8 & 20.8 & 52.3 & 59.8 & 70.0 & 59.5 \\ \hline
\end{tabular}
\end{adjustbox}
\caption{Ablation comparison of BEV feature pooling strategies.}
\label{tab:pool}
\end{table}

\paragraph{\textbf{BEV Feature Pooling Strategy}}
In terms of the feature pooling strategy for the cropped regions, we compared the classic Max Pooling and Mean Pooling operations. As illustrated in Table \ref{tab:pool}, Max Pooling achieved an accuracy of $58.9\%$ under the same conditions, which is $0.6\%$ lower than Mean Pooling. We speculate that this difference may be due to the fact that Max Pooling focuses on the texture features within the region, while Mean Pooling preserves the overall features. Our proposed NuScenes-QA mainly tests the model's ability of understanding the structure of objects and their spatial relationships in street views, and relatively ignores the texture of the objects. Thus, Mean Pooling has a slight advantage over Max Pooling.

\section{Conclusion}

In this paper, we apply VQA to the context of autonomous driving. We construct NuScenes-QA, the first large-scale multi-modal VQA benchmark for autonomous driving scenario. NuScenes-QA are generated automatically based on visual scene graphs and question templates, containing 34K scenes and 460K question-answer pairs. Alongside a series of baseline models, comprehensive experiments establish a solid foundation for future research. We strongly hope that NuScenes-QA can invigorate the evolution of multi-modal VQA and propel advancements in autonomous driving.

\section{Acknowledgments}
This work was supported in part by National Natural Science Foundation of China Project (No. 62072116) and Shanghai Science and Technology Program [Project No. 21JC1400600].

\bibliography{aaai24}

\newpage
\appendix
\part*{Appendix}

\begin{table*}
    \centering
    \begin{tabular}{ccccccc}
    \hline
Dataset & Visual Modality & Multi Frame & Scenario & Collection & \# Scenes & \#Amount \\ \hline
EQA \cite{das2018embodied} & image & \Checkmark & indoor & template & 767 & 1.5k \\ 
3DQA \cite{ye20223d} & point cloud & \XSolidBrush & indoor & human & 806 & 10k \\
ScanQA \cite{azuma2022scanqa} & point cloud & \XSolidBrush & indoor & auto + human & 800 & 41k \\
CLEVR3D \cite{yan2021clevr3d} & point cloud & \XSolidBrush & indoor & template & 1,129 & 60.1k \\
FE-3DGQA \cite{zhao2022towards} & point cloud & \XSolidBrush & indoor & human & 800 & 20k \\
SQA3D \cite{ma2022sqa3d} & image + point cloud & \Checkmark & indoor & human & 650 & 33.4k \\ \hline 
NuScenes-QA (ours) & image + point cloud & \Checkmark & outdoor & template & 34,149 & 460k \\ \hline
\end{tabular}
    \caption{Comparison between NuScenes-QA and other VQA datasets.}
    \label{tab:com}
\end{table*}

\section{Dataset Comparison}
\label{sec:com}

Table \ref{tab:com} summarizes the differences between our NuScenes-QA and other related 3D VQA datasets. Firstly, to the best of our knowledge, NuScenes-QA is currently the largest 3D VQA datasets, with 34k diverse visual scenes and 460k question-answer pairs, averaging 13.5 pairs per scene. In contrast, other datasets \cite{das2018embodied, ye20223d, azuma2022scanqa, yan2021clevr3d, zhao2022towards, ma2022sqa3d} typically have less than 1000 scenes due to the difficulty of acquiring 3D data. Secondly, in terms of visual modality, our NuScenes-QA is multimodal, comprising of images and point clouds, posing higher demands on visual reasoning while increasing the dataset's complexity. Additionally, our visual data is multi-frame, requiring temporal information mining. Thirdly, Our dataset focuses on outdoor scenes with a combination of dynamic objects and static background, posing higher challenges for answering questions that require perceiving and reasoning about dynamic objects.

In summary, compared to other 3D VQA datasets, NuScenes-QA stands out in terms of scale, data modality, and content, making it an important research resource that can advance the research and development of 3D visual question answering.

\section{Question Templates}
\label{sec:tem}

As described in the main paper, we employ manually designed question templates to generate questions programmatically. For instance, the question ``\textit{Are there any bicycles to the back left of the moving car?}'' has a template of ``\textit{Are there any \textless A2\textgreater \,\textless O2\textgreater s to the \textless R\textgreater \,of the \textless A\textgreater \,\textless O\textgreater?}'', where \textit{\textless A\textgreater}, \textit{\textless O\textgreater}, and \textit{\textless R\textgreater} denote status, object, and relation, respectively. Table \ref{tab:tem} enumerates all the templates based on different question types. Among them, one-hop questions involve reasoning about relations between objects, while zero-hop questions are relatively simpler. We design a total of 5 question types and 16 different semantic question templates. To increase question diversity, we created different variations of each template, such as ``\textit{How many \textless A2\textgreater \,\textless O2\textgreater s are to the \textless R\textgreater \,of the \textless A\textgreater \,\textless O\textgreater?}'' can also be expressed as ``\textit{There is a \textless A\textgreater \,\textless O\textgreater; how many \textless A2\textgreater \,\textless O2\textgreater s are to the \textless R\textgreater \,of it?}'' In the end, we get a total of 66 different question templates. In future work, we can further enrich the question templates to enhance the diversity of questions.

\begin{table*}
\begin{adjustbox}{width=\textwidth}
\begin{tabular}{|c|c|l|}
\hline
Type                                                                    & Number of Hop      & \multicolumn{1}{c|}{Template}\\ \hline
\multirow{11}{*}{Existence}                                              & \multirow{5}{*}{0} & \begin{tabular}[c]{@{}l@{}}Are there any \textless{}A\textgreater \textless{}O\textgreater{}s?\\ Are any \textless{}A\textgreater \textless{}O\textgreater{}s visible?\end{tabular}\\ \cline{3-3} 
& & \begin{tabular}[c]{@{}l@{}}Are there any other \textless{}A2\textgreater \textless{}O2\textgreater{}s that in the same status as the \textless{}A\textgreater \textless{}O\textgreater{}?\\ Is there another \textless{}A2\textgreater \textless{}O2\textgreater that has the same status as the \textless{}A\textgreater \textless{}O\textgreater{}?\\ Are there any other \textless{}A2\textgreater \textless{}O2\textgreater{}s of the same status as the \textless{}A\textgreater \textless{}O\textgreater{}?\\ Is there another \textless{}A2\textgreater \textless{}O2\textgreater of the same status as the \textless{}A\textgreater \textless{}O\textgreater{}?\end{tabular}\\ \cline{2-3} 
& \multirow{5}{*}{1} & \begin{tabular}[c]{@{}l@{}}Are there any \textless{}A2\textgreater \textless{}O2\textgreater{}s to the \textless{}R\textgreater of the \textless{}A\textgreater \textless{}O\textgreater{}?\\ There is a \textless{}A\textgreater \textless{}O\textgreater{}; are there any \textless{}A2\textgreater \textless{}O2\textgreater{}s to the \textless{}R\textgreater of it?\end{tabular}\\ \cline{3-3} 
& & \begin{tabular}[c]{@{}l@{}}Are there any other \textless{}A3\textgreater \textless{}O3\textgreater{}s that in the same status as the \textless{}A2\textgreater \textless{}O2\textgreater {[}that is{]} to the \textless{}R\textgreater of the \textless{}A\textgreater \textless{}O\textgreater{}?\\ Is there another \textless{}A3\textgreater \textless{}O3\textgreater that has the same status as the \textless{}A2\textgreater \textless{}O2\textgreater {[}that is{]} to the \textless{}R\textgreater of the \textless{}A\textgreater \textless{}O\textgreater{}?\\ Are there any other \textless{}A3\textgreater \textless{}O3\textgreater{}s of the same status as the \textless{}A2\textgreater \textless{}O2\textgreater {[}that is{]} to the \textless{}R\textgreater of the \textless{}A\textgreater \textless{}O\textgreater{}?\\ Is there another \textless{}A3\textgreater \textless{}O3\textgreater of the same status as the \textless{}A2\textgreater \textless{}O2\textgreater {[}that is{]} to the \textless{}R\textgreater of the \textless{}A\textgreater \textless{}O\textgreater{}?\end{tabular}\\ \hline

\multirow{12}{*}{Counting}                                               & \multirow{5}{*}{0} & \begin{tabular}[c]{@{}l@{}}How many \textless{}A\textgreater \textless{}O\textgreater{}s are there?\\ What number of \textless{}A\textgreater \textless{}O\textgreater{}s are there?\end{tabular}\\ \cline{3-3} 
& & \begin{tabular}[c]{@{}l@{}}How many other things are in the same status as the \textless{}A\textgreater \textless{}O\textgreater{}?\\ What number of other things are in the same status as the \textless{}A\textgreater \textless{}O\textgreater{}?\\ How many other things are there of the same status as the \textless{}A\textgreater \textless{}O\textgreater{}?\\ What number of other things are there of the same status as the \textless{}A\textgreater \textless{}O\textgreater{}?\end{tabular}\\ \cline{2-3} 
& \multirow{4}{*}{1} & \begin{tabular}[c]{@{}l@{}}What number of \textless{}A2\textgreater \textless{}O2\textgreater{}s are to the \textless{}R\textgreater of the \textless{}A\textgreater \textless{}O\textgreater{}?\\ How many \textless{}A2\textgreater \textless{}O2\textgreater{}s are to the \textless{}R\textgreater of the \textless{}A\textgreater \textless{}O\textgreater{}?\\ There is a \textless{}A\textgreater \textless{}O\textgreater{}; how many \textless{}A2\textgreater \textless{}O2\textgreater{}s are to the \textless{}R\textgreater of it?\\ There is a \textless{}A\textgreater \textless{}O\textgreater{}; what number of \textless{}A2\textgreater \textless{}O2\textgreater{}s are to the \textless{}R\textgreater of it?\end{tabular}\\ \cline{3-3} 
& & \begin{tabular}[c]{@{}l@{}}How many other \textless{}A3\textgreater \textless{}O3\textgreater{}s in the same status as the \textless{}A2\textgreater \textless{}O2\textgreater {[}that is{]} to the \textless{}R\textgreater of the \textless{}A\textgreater \textless{}O\textgreater{}?\\ How many other \textless{}A3\textgreater \textless{}O3\textgreater{}s are in the same status as the \textless{}A2\textgreater \textless{}O2\textgreater {[}that is{]} to the \textless{}R\textgreater of the \textless{}A\textgreater \textless{}O\textgreater{}?\\ What number of other \textless{}A3\textgreater \textless{}O3\textgreater{}s in the same status as the \textless{}A2\textgreater \textless{}O2\textgreater {[}that is{]} to the \textless{}R\textgreater of the \textless{}A\textgreater \textless{}O\textgreater{}?\end{tabular}\\ \hline

\multirow{7}{*}{\begin{tabular}[c]{@{}c@{}}Query\\ object\end{tabular}} & 0 & \begin{tabular}[c]{@{}l@{}}What is the \textless{}A\textgreater \textless{}O\textgreater{}?\\ The \textless{}A\textgreater \textless{}O\textgreater is what?\\ There is a \textless{}A\textgreater \textless{}O\textgreater{}; what is it?\end{tabular}\\ \cline{2-3} 
& \multirow{4}{*}{1} & \begin{tabular}[c]{@{}l@{}}What is the \textless{}A2\textgreater \textless{}O2\textgreater {[}that is{]} to the \textless{}R\textgreater of the \textless{}A\textgreater \textless{}O\textgreater{}?\\ The \textless{}A2\textgreater \textless{}O2\textgreater {[}that is{]} to the \textless{}R\textgreater of the \textless{}A\textgreater \textless{}O\textgreater is what?\\ There is a \textless{}A2\textgreater \textless{}O2\textgreater {[}that is{]} to the \textless{}R\textgreater of the \textless{}A\textgreater \textless{}O\textgreater{}; what is it?\end{tabular}\\ \cline{3-3} 
& & \begin{tabular}[c]{@{}l@{}}What is the \textless{}A3\textgreater \textless{}O3\textgreater that is {[}both{]} to the \textless{}R2\textgreater of the \textless{}A2\textgreater \textless{}O2\textgreater and the \textless{}R\textgreater of the \textless{}A\textgreater \textless{}O\textgreater{}?\\ The \textless{}A3\textgreater \textless{}O3\textgreater that is {[}both{]} to the \textless{}R2\textgreater of the \textless{}A2\textgreater \textless{}O2\textgreater and the \textless{}R\textgreater of the \textless{}A\textgreater \textless{}O\textgreater is what?\\ There is a \textless{}A3\textgreater \textless{}O3\textgreater that is {[}both{]} to the \textless{}R2\textgreater of the \textless{}A2\textgreater \textless{}O2\textgreater and the \textless{}R\textgreater of the \textless{}A\textgreater \textless{}O\textgreater{}; what is it?\end{tabular}\\ \hline

\multirow{6}{*}{\begin{tabular}[c]{@{}c@{}}Query\\ status\end{tabular}} & 0                  & \begin{tabular}[c]{@{}l@{}}What status is the \textless{}A\textgreater \textless{}O\textgreater{}?\\ What is the status of the \textless{}A\textgreater \textless{}O\textgreater{}?\\ The \textless{}A\textgreater \textless{}O\textgreater is in what status?\\ There is a \textless{}A\textgreater \textless{}O\textgreater{}; what status is it?\end{tabular}\\ \cline{2-3} 
& 1                  & \begin{tabular}[c]{@{}l@{}}What status is the \textless{}A2\textgreater \textless{}O2\textgreater {[}that is{]} to the \textless{}R\textgreater of the \textless{}A\textgreater \textless{}O\textgreater{}?\\ What is the status of the \textless{}A2\textgreater \textless{}O2\textgreater {[}that is{]} to the \textless{}R\textgreater of the \textless{}A\textgreater \textless{}O\textgreater{}?\\ The \textless{}A2\textgreater \textless{}O2\textgreater {[}that is{]} to the \textless{}R\textgreater of the \textless{}A\textgreater \textless{}O\textgreater is in what status?\\ There is a \textless{}A2\textgreater \textless{}O2\textgreater {[}that is{]} to the \textless{}R\textgreater of the \textless{}A\textgreater \textless{}O\textgreater{}; what status is it?\\ There is a \textless{}A2\textgreater \textless{}O2\textgreater {[}that is{]} to the \textless{}R\textgreater of the \textless{}A\textgreater \textless{}O\textgreater{}; what is its status?\end{tabular}\\ \hline

\multirow{18}{*}{Comparison}                                             & 0                  & \begin{tabular}[c]{@{}l@{}}Do the \textless{}A\textgreater \textless{}O\textgreater and the \textless{}A2\textgreater \textless{}O2\textgreater have the same status?\\ Is the status of the \textless{}A\textgreater \textless{}O\textgreater the same as the \textless{}A2\textgreater \textless{}O2\textgreater{}?\\ Do the \textless{}A\textgreater \textless{}O\textgreater and the \textless{}A2\textgreater \textless{}O2\textgreater have the same status?\\ Is the \textless{}A\textgreater \textless{}O\textgreater the same status as the \textless{}A2\textgreater \textless{}O2\textgreater{}?\\ Does the \textless{}A\textgreater \textless{}O\textgreater have the same status as the \textless{}A2\textgreater \textless{}O2\textgreater{}?\end{tabular}\\ \cline{2-3} 
& \multirow{13}{*}{1} & \begin{tabular}[c]{@{}l@{}}Is the status of the \textless{}A2\textgreater \textless{}O2\textgreater {[}that is{]} to the \textless{}R\textgreater of the \textless{}A\textgreater \textless{}O\textgreater the same as the \textless{}A3\textgreater \textless{}O3\textgreater{}?\\ Does the \textless{}A2\textgreater \textless{}O2\textgreater {[}that is{]} to the \textless{}R\textgreater of the \textless{}A\textgreater \textless{}O\textgreater have the same status as the \textless{}A3\textgreater \textless{}O3\textgreater{}?\\ Do the \textless{}A2\textgreater \textless{}O2\textgreater {[}that is{]} to the \textless{}R\textgreater of the \textless{}A\textgreater \textless{}O\textgreater and the \textless{}A3\textgreater \textless{}O3\textgreater have the same status?\\ There is a \textless{}A2\textgreater \textless{}O2\textgreater {[}that is{]} to the \textless{}R\textgreater of the \textless{}A\textgreater \textless{}O\textgreater{}; does it have the same status as the \textless{}A3\textgreater \textless{}O3\textgreater{}?\\ There is a \textless{}A2\textgreater \textless{}O2\textgreater {[}that is{]} to the \textless{}R\textgreater of the \textless{}A\textgreater \textless{}O\textgreater{}; is it the same status as the \textless{}A3\textgreater \textless{}O3\textgreater{}?\\ There is a \textless{}A2\textgreater \textless{}O2\textgreater {[}that is{]} to the \textless{}R\textgreater of the \textless{}A\textgreater \textless{}O\textgreater{}; is its status the same as the \textless{}A3\textgreater \textless{}O3\textgreater{}?\end{tabular}\\ \cline{3-3} 
&                    & \begin{tabular}[c]{@{}l@{}}Is the status of the \textless{}A\textgreater \textless{}O\textgreater the same as the \textless{}A3\textgreater \textless{}O3\textgreater {[}that is{]} to the \textless{}R\textgreater of the \textless{}A2\textgreater \textless{}O2\textgreater{}?\\ Does the \textless{}A\textgreater \textless{}O\textgreater have the same status as the \textless{}A3\textgreater \textless{}O3\textgreater {[}that is{]} to the \textless{}R\textgreater of the \textless{}A2\textgreater \textless{}O2\textgreater{}?\\ Do the \textless{}A\textgreater \textless{}O\textgreater and the \textless{}A3\textgreater \textless{}O3\textgreater {[}that is{]} to the \textless{}R\textgreater of the \textless{}A2\textgreater \textless{}O2\textgreater have the same status?\\ There is a \textless{}A\textgreater \textless{}O\textgreater{}; does it have the same status as the \textless{}A3\textgreater \textless{}O3\textgreater {[}that is{]} to the \textless{}R\textgreater of the \textless{}A2\textgreater \textless{}O2\textgreater{}?\\ There is a \textless{}A\textgreater \textless{}O\textgreater{}; is its status the same as the \textless{}A3\textgreater \textless{}O3\textgreater {[}that is{]} to the \textless{}R\textgreater of the \textless{}A2\textgreater \textless{}O2\textgreater{}?\\ There is a \textless{}A\textgreater \textless{}O\textgreater{}; is it the same status as the \textless{}A3\textgreater \textless{}O3\textgreater {[}that is{]} to the \textless{}R\textgreater of the \textless{}A2\textgreater \textless{}O2\textgreater{}?\end{tabular}\\ \cline{3-3} 
&                    & \begin{tabular}[c]{@{}l@{}}Is the status of the \textless{}A2\textgreater \textless{}O2\textgreater {[}that is{]} to the \textless{}R\textgreater of the \textless{}A\textgreater \textless{}O\textgreater the same as the \textless{}A4\textgreater \textless{}O4\textgreater {[}that is{]} to the \textless{}R2\textgreater of the \textless{}A3\textgreater \textless{}O3\textgreater{}?\\ Does the \textless{}A2\textgreater \textless{}O2\textgreater {[}that is{]} to the \textless{}R\textgreater of the \textless{}A\textgreater \textless{}O\textgreater have the same status as the \textless{}A4\textgreater \textless{}O4\textgreater {[}that is{]} to the \textless{}R2\textgreater of the \textless{}A3\textgreater \textless{}O3\textgreater{}?\\ Do the \textless{}A2\textgreater \textless{}O2\textgreater {[}that is{]} to the \textless{}R\textgreater of the \textless{}A\textgreater \textless{}O\textgreater and the \textless{}A4\textgreater \textless{}O4\textgreater {[}that is{]} to the \textless{}R2\textgreater of the \textless{}A3\textgreater \textless{}O3\textgreater have the same status?\\ There is a \textless{}A2\textgreater \textless{}O2\textgreater {[}that is{]} to the \textless{}R\textgreater of the \textless{}A\textgreater \textless{}O\textgreater{}; does it have the same status as the \textless{}A4\textgreater \textless{}O4\textgreater {[}that is{]} to the \textless{}R2\textgreater of the \textless{}A3\textgreater \textless{}O3\textgreater{}?\\ There is a \textless{}A2\textgreater \textless{}O2\textgreater {[}that is{]} to the \textless{}R\textgreater of the \textless{}A\textgreater \textless{}O\textgreater{}; is its status the same as the \textless{}A4\textgreater \textless{}O4\textgreater {[}that is{]} to the \textless{}R2\textgreater of the \textless{}A3\textgreater \textless{}O3\textgreater{}?\\ There is a \textless{}A2\textgreater \textless{}O2\textgreater {[}that is{]} to the \textless{}R\textgreater of the \textless{}A\textgreater \textless{}O\textgreater{}; is it the same status as the \textless{}A4\textgreater \textless{}O4\textgreater {[}that is{]} to the \textless{}R2\textgreater of the \textless{}A3\textgreater \textless{}O3\textgreater{}?\end{tabular} \\ \hline
\end{tabular}
\end{adjustbox}
\caption{Question templates.}
\label{tab:tem}
\end{table*}

\begin{figure}
    \centering
    \includegraphics[width=0.4\textwidth]{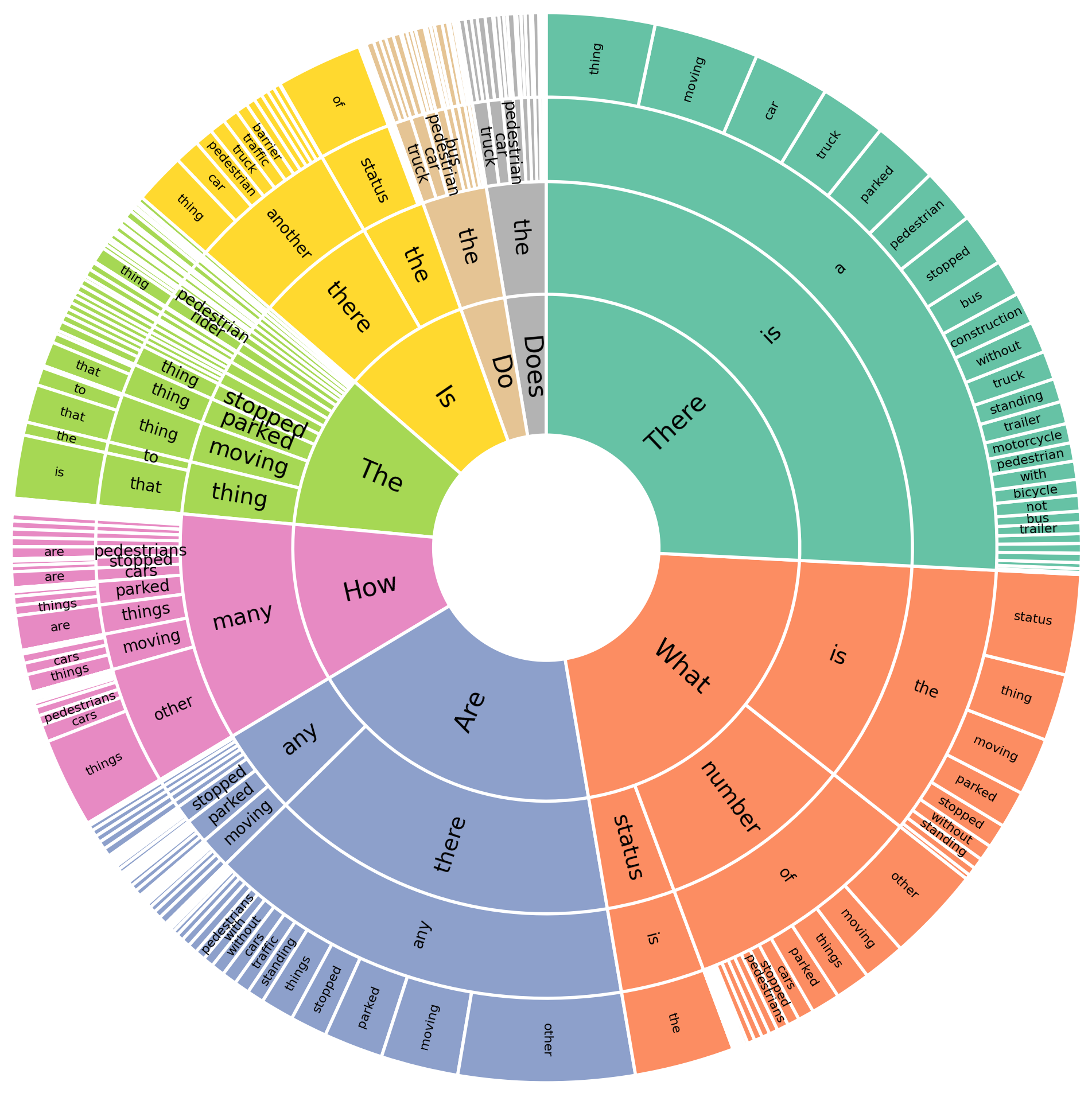}
    \caption{Question distribution in terms of the first four words.}
    \label{fig:four_word}
\end{figure}

\section{Question Distribution of the First Four Words}

In Figure \ref{fig:four_word}, we visualize the distribution of the first four words in the questions, from which we can draw two observations. First, our data distribution is balanced, as already verified in Figure \ref{fig:status}. Second, our questions encompass a diverse range of visual semantics. To answer them, not only the object categories such as pedestrian, motorcycle are demanded, but also their status, such as moving or parked. The semantic complexities in our questions also presents a considerable challenge for models.

\section{Visualization Examples}
\label{sec:vis}

\begin{figure*}
    \centering
    \includegraphics[width=0.9\textwidth]{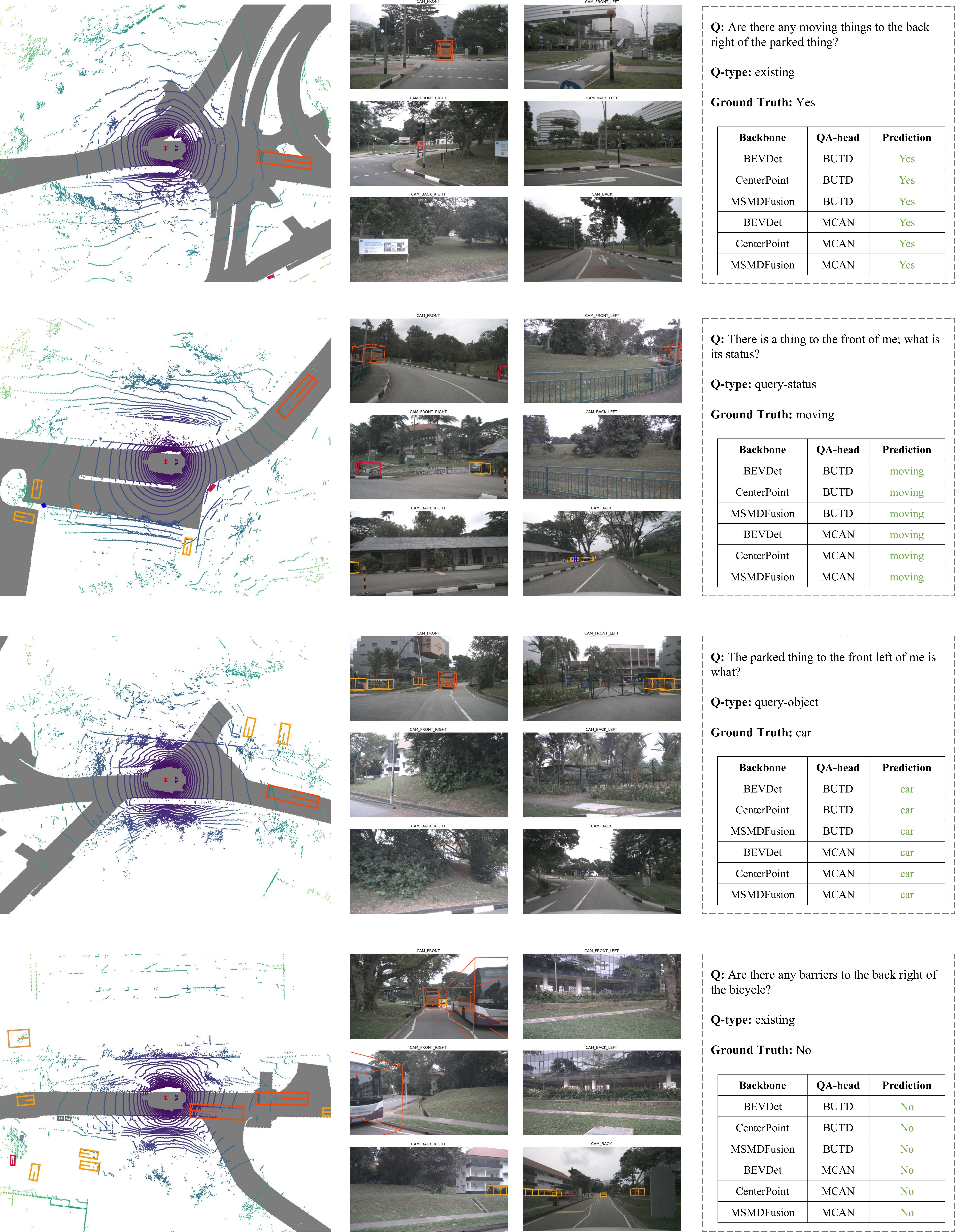}
    \caption{Some good cases.}
    \label{fig:visualization_1}
\end{figure*}

\begin{figure*}
    \centering
    \includegraphics[width=0.9\textwidth]{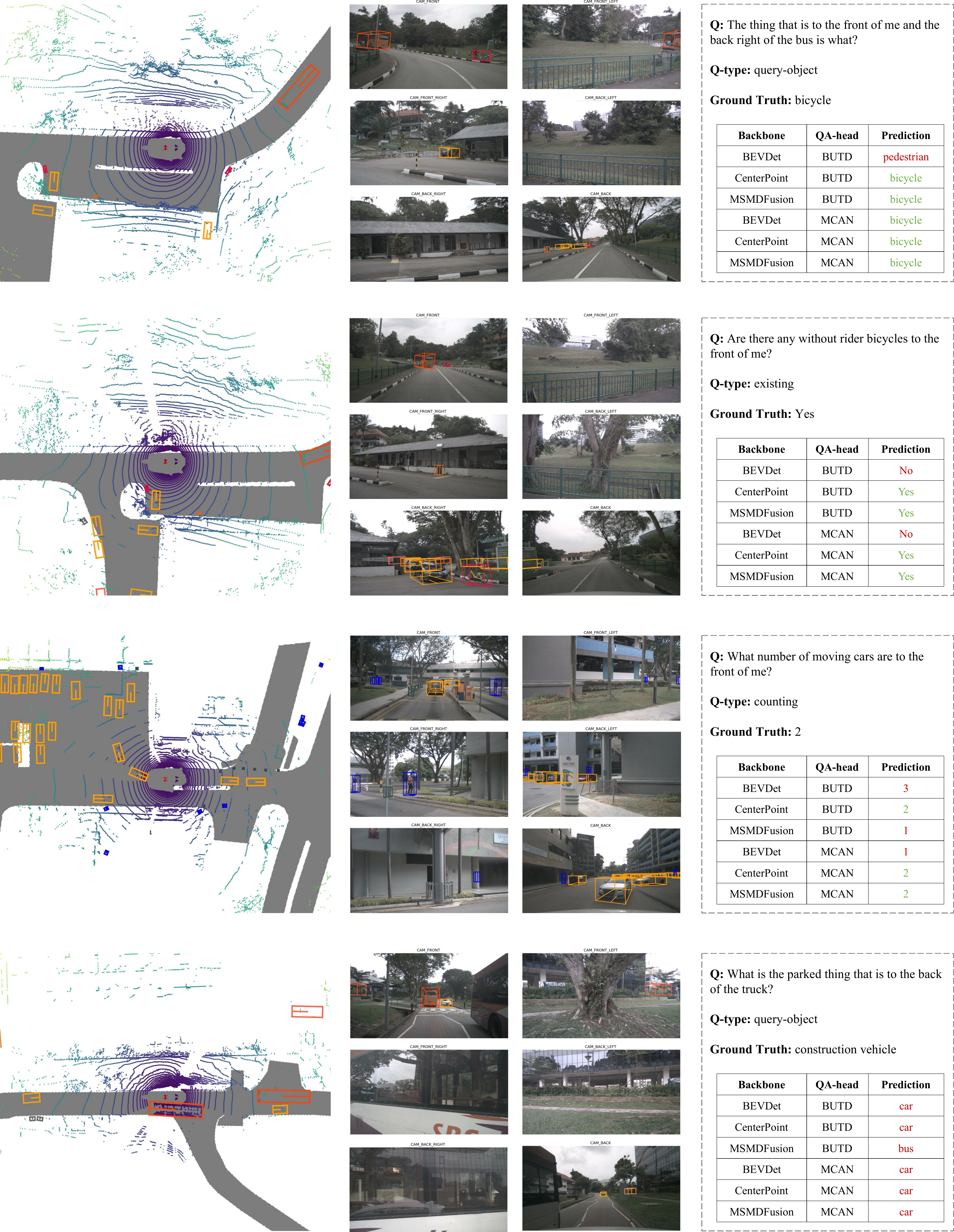}
    \caption{Some bad cases.}
    \label{fig:visualization_2}
\end{figure*}

In order to get a deeper insights of the difficulty of the NuScenes-QA dataset and to validate the performance of proposed baseline models, we select some samples from the test split for visualization, as shown in Figure \ref{fig:visualization_1} and Figure \ref{fig:visualization_2}. We present the point clouds, surround-view images, questions, ground truth answers, and the predictions of 6 different baseline models.

The visual data of these examples (with LiDAR point clouds on the left, surround-view images in the middle) provide compelling evidence of the challenges inherent in NuScenes-QA. First, the point clouds and images are in different coordinate systems, which poses great difficulties for the model to comprehend spatial relationships and fuse multi-modal information. Second, the street-view scenes often contain a lot of background noise, which can significantly interfere with the model's ability to locate and recognize foreground objects relevant to the question. Thirdly, some scenes contain dense and diverse objects, making the reasoning for questions in these complex scenes prone to errors.

Figure \ref{fig:visualization_1} shows some successful cases, demonstrating the impressive performance of the baseline models. Inevitably, there are also failed cases, as shown in Figure \ref{fig:visualization_2}. By comparing the prediction results of different baselines, we find that multi-modal fusion models usually outperform single-modal models, which confirms the complementarity between point clouds and images.

\section{Discussion}
\paragraph{\textbf{Limitations}}

As the first work in this field, there are still many shortcomings in NuScenes-QA that need to be improved. First, our questions focus on foreground objects and their relationships, but ignore the connection between objects and the background. For example, we can ask the agent which parking spots ahead are available, which is also important for autonomous driving. Second, our questions focus on the model's understanding of the visual scene, which may be too naive for humans. Meanwhile, traditional image or video based question answering is gradually moving towards higher levels of reasoning such as causal inference. Third, we define the relationships between objects using spatial positioning only, lacking many rich semantic relationships. Finally, although NuScenes-QA is large scale, our questions may not be sufficiently diverse in language due to templates based generation. In summary, there is still significant room for improvement in NuScenes-QA.

\paragraph{\textbf{Future Work}}
In future work, we can gradually improve NuScenes-QA in the following aspects. Firstly, we can add object localization tasks, like ScanQA \cite{azuma2022scanqa}, to increase the model's interpretability. Secondly, we can utilize crowd sourcing to manually annotate the scene graphs, including new relationships, object status, and more. This can enrich the semantics of questions and increase the diversity of language. Lastly, we can consider integrating with perception tasks such as tracking to increase its practical value.

\end{document}